\documentclass[a4paper]{article}
\usepackage{Odyssey2024}
\usepackage{epsfig,amssymb,amsmath}
\usepackage{booktabs} 
\usepackage{multirow} 
\usepackage{cite} 
\usepackage{tabularx}
\usepackage{stackengine}
\usepackage{url}
\usepackage{caption} 
\usepackage{xcolor}
\usepackage{pifont}
 \usepackage{threeparttable}
\definecolor{myred}{RGB}{0,0,0}
\ninept

\title{Improving Speaker Assignment in Speaker-Attributed ASR\\for Real Meeting Applications}

\makeatletter
\def\name#1{\gdef\@name{#1\\}}
\makeatother
\name{Can Cui$^{1, 2}$, Imran Sheikh$^2$, Mostafa Sadeghi$^1$, Emmanuel Vincent$^1$}
\address{$^1$Université de Lorraine, CNRS, Inria, LORIA, F-54000 Nancy, France\\
$^2$Vivoka, Metz, France \\
{\small \tt $^1$firstname.lastname@inria.fr,$^2$firstname.lastname@vivoka.com } }

\begin{document}
\maketitle

\begin{abstract}
Past studies on end-to-end meeting transcription have focused on model architecture and have mostly been evaluated on simulated meeting data. We present a novel study aiming to optimize the use of a Speaker-Attributed ASR (SA-ASR) system in real-life scenarios, such as the AMI meeting corpus, for improved speaker assignment of speech segments. First, we propose a pipeline tailored to real-life applications involving Voice Activity Detection (VAD), Speaker Diarization (SD), and SA-ASR. Second, we advocate using VAD output segments to fine-tune the SA-ASR model, considering that it is also applied to VAD segments during test, and show that this results in a relative reduction of Speaker Error Rate (SER) up to 28\%. Finally, we explore strategies to enhance the extraction of the reference speaker embeddings used as inputs by the SA-ASR system. We show that extracting them from SD output rather than annotated speaker segments results in a relative SER reduction up to \textcolor{myred} {16\%}. 

\end{abstract}

\section{Introduction}
\label{sec:intro}

Multi-speaker meeting transcription has recently become a vibrant and active research topic
\cite{yu2022m2met, yoshioka2019advances,yoshioka2019meeting,cornell2023chime}. This task poses various challenges, including distant microphones, overlapped speech, and ambient noise. The earliest works on end-to-end (E2E) models for this task \cite{9054029,seki-etal-2018-purely} had demonstrated the advantages of an E2E model over a traditional system wherein different components address different subtasks.
Recent research primarily concentrates on enhancing the E2E model architectures \cite{kanda2021end, yu2022comparative,chang2021hypothesis}. The work in \cite{kanda2021end} introduced a Transformer-based E2E Speaker-Attributed ASR (SA-ASR) system for simultaneous recognition of speech and speaker identities. Thereafter, SA-ASR was extended to accommodate diverse scenarios \cite{kanda2021investigation}, address an unlimited number of speakers \cite{kanda2022transcribe}, and cater to streaming applications \cite{kanda2022streaming}. Moreover, non-autoregressive models for multi-speaker ASR \cite{guo2021multi} have been proposed.

While the model architecture is undoubtedly crucial, 
the application of the E2E model to real-life data has not received much attention. 
Due to the lack of significant amounts of real meeting-style training data, training of multi-speaker ASR models relies on simulated multi-speaker overlapped distant speech data. To improve performance on real test data, these models are adapted or fine-tuned on real training data.
A few studies \cite{kanda2021large, cui2023, yang2023simulating} have discussed training and evaluation on real recorded multi-speaker datasets, such as the AMI meeting corpus \cite{carletta2005ami} which still remains a challenging multi-speaker speech transcription task.

Training and inference on real long-length multi-speaker audio recordings is a non-trivial issue due to computational and memory requirements.
One approach to address this issue is to segment the long recording at
silence positions into disjoint segments \cite{kanda2021large}. Another approach is to segment it into fixed-sized chunks and adjust the start/end times to non-overlapped word boundaries \cite{cui2023}. Both approaches rely on ground-truth
\textcolor{myred}{annotations of the test set}. In real-life applications, this information is not available and a Voice Activity Detection (VAD) system becomes essential for obtaining speech segments.

A second problem with training and inference on real multi-speaker audio is how to obtain the reference speaker embeddings \textcolor{myred}{(a.k.a.\ speaker profiles)} used as inputs by the SA-ASR system.
In real-life applications, front-end Speaker Diarization (SD) becomes essential for 
this purpose. While studies on VAD \cite{medennikov2020target,yang2010comparative} and SD \cite{park2022review,xiao2021microsoft} exist independently, there is little discussion on how to optimally integrate them into an SA-ASR pipeline to create a ready-to-use meeting transcription system.
Speaker assignment in SA-ASR is typically performed using an x-vector-based speaker embedding model \cite{snyder2018x}. 
Preparation of a representative template for each speaker typically involves averaging the embeddings of candidate segments \cite{kanda2021end,yu2022comparative,cui2023}. However, the selection of segments for better representing each speaker remains a question. 

The contributions of this paper include: 
(a) a VAD-SD-SA-ASR pipeline for real meeting transcription; 
(b) fine-tuning of SA-ASR on VAD output segments instead of ground-truth or fixed-sized segments to better fit the test conditions; 
(c) a discussion on 
strategies to improve speaker assignment in SA-ASR by leveraging various attributes related to VAD and SD, such as the number and length of segments used to obtain speaker profiles.
Experiments show that our proposed fine-tuning method reduces the SER by up to 28\% relative and that extracting speaker profiles from SD output rather than annotated speaker segments results further reduces it by up to \textcolor{myred} {16\% }relative.
To the best of our knowledge, this is the first study that investigates how attributes such as the number and the length of segments impact the accuracy of the speaker profiles and the resulting SA-ASR performance.

The rest of the paper is organized as follows. Section~\ref{sec:related} presents the individual modules in our pipeline, including the segmentation methods for raw meeting audio. Section~\ref{sec:methods} describes our pipeline approach and the preparation of the training, development, and test data. Section~\ref{sec:exp} presents our experimental settings, followed by the discussion of results in Section~\ref{sec:discuss}. Finally, Section~\ref{sec:conclusion} provides a conclusion.
 
\section{Background}
\label{sec:related}
\subsection{Voice Activity Detection: CRDNN}
\label{subsec:vad}
A Voice Activity Detection (VAD) system identifies speech and non-speech segments in an audio signal, which is crucial for applications like speech recognition. Neural network-based models for VAD often include architectures like Convolutional Neural Networks \cite{asif2022emotion}, Recurrent Neural Networks \cite{lin202347}, Long Short-Term Memory networks \cite{kim2016vowel}, etc. One of the common architectures is CRDNN (Convolutional, Recurrent, and Dense Neural Network) \cite{sainath2015convolutional, xiang2021fast}. It combines convolutional layers for frequency variations, recurrent layers for temporal modeling, and dense layers for feature mapping, which have succeeded in various speech-related tasks. 

\subsection{Speaker Diarization: ECAPA-TDNN}
\label{subsec:ecapa} 
Classical Speaker Diarization (SD) systems commonly rely on clustering speaker embeddings \cite{park2022review} such as i-vectors \cite{dehak2010front}, d-vectors \cite{variani2014deep}, and x-vectors \cite{snyder2018x}. Specifically, ECAPA-TDNN (Emphasized Channel Attention, Propagation, and Aggregation in Time-Delay Neural Network) \cite{desplanques2020ecapa} is a model used for computing x-vectors, which are widely employed for speaker representation in SD tasks.
It enhances the traditional Time-Delay Neural Network (TDNN) architecture by integrating channel attention mechanisms, which dynamically emphasize relevant information and suppress noise, resulting in more robust speaker embeddings. ECAPA-TDNN utilizes both forward and backward information propagation to capture long-term dependencies and context in input sequences, ensuring effective modeling of temporal relationships for speaker verification. Additionally, it employs aggregation techniques to combine information across multiple frames, creating a comprehensive representation of speaker characteristics. \textcolor{myred}{Our SD system utilizes Spectral Clustering \cite{ning2006spectral} as clustering algorithm, and the overlapped speech segments are split equally among the adjacent segments with different speakers.}

\begin{figure}[b!]
  \centering
  \includegraphics[width=1\linewidth]{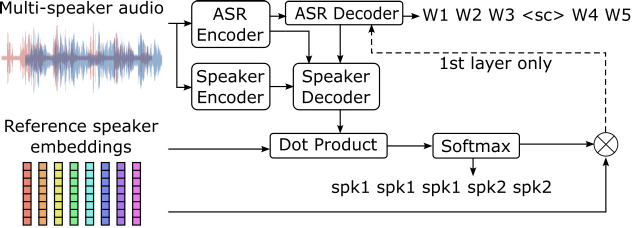}
  \caption{\textcolor{myred}{Diagram of SA-ASR \cite{kanda2021end}}.}
  \label{fig:saasr}
\end{figure}

\subsection{End-to-end Speaker-Attributed ASR}
\label{subsec:saasr}

A Transformer-based end-to-end Speaker-Attributed ASR (SA-ASR) system was proposed in \cite{kanda2021end}. Following the Serialized Output Training (SOT) principle \cite{kanda2020serialized}, the output is the concatenation of all speakers' sentences in first-in-first-out order, where each token is associated with one speaker ID and distinct speakers are separated by an \textless sc\textgreater\ token. As shown in Figure~\ref{fig:saasr}, the inputs to the model consist of an acoustic feature sequence and a set of reference speaker embeddings 
obtained from enrollment data. A Conformer-based ASR Encoder first encodes acoustic information, along with a Speaker Encoder to encode speaker information. Subsequently, the Transformer-based ASR Decoder and Speaker Decoder modules decode text and speaker information, respectively. The ASR block and the speaker block mutually share information through the ASR Decoder and the Speaker Decoder, enabling mutual assistance between text recognition and speaker identification. In addition to the ASR decoder's output, which consists of tokens corresponding to different speakers, the Speaker Decoder computes a speaker representation 
corresponding to each ASR output token.
This representation is used to assign speakers 
by computing a dot product with the speaker embedding
templates.

\begin{figure}[b!]
  \centering
  \includegraphics[width=1\linewidth]{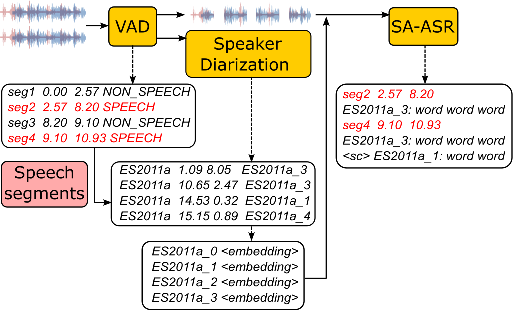}
  \caption{\textcolor{myred}{Pipeline of the proposed system.}}
  \label{fig:system}
\end{figure}

\subsection{AMI segmentation}
\label{subsec:ami-seg}

\subsubsection{Based on fixed-sized chunks}
\label{subsubsec:fixed}
The AMI corpus has been recently segmented using fixed chunk and hop sizes \cite{cui2023}. If the start/end time of a segment is within an overlapped speech region, it is modified to be 2~s outside of the overlap region. If the start/end time of a segment falls within a word, it is adjusted to the start/end time of that word using AMI's word boundary annotations.
This approach simulates how to process meeting audio in situations where ground-truth silence/non-speech positions between utterances are not available at test time. However, speaker overlap and word boundary annotations are still required.

\subsubsection{Based on ground-truth silence positions}
\label{subsubsec:gold}

Multi-speaker ASR systems applied on the AMI corpus typically use ground-truth segment annotations for fine-tuning and test \cite{kanda2021large, yang2023simulating}. More specifically, in \cite{kanda2021large}, the concept of an ``utterance group''  is introduced by segmenting the recording at silence/non-speech positions between utterances. This allows the model to transcribe text and speaker information more effectively using the first-in-first-out principle.
 
\section{Proposed methods}
\label{sec:methods}

\begin{figure*}[t]
  \centering
  \includegraphics[width=0.9\linewidth]{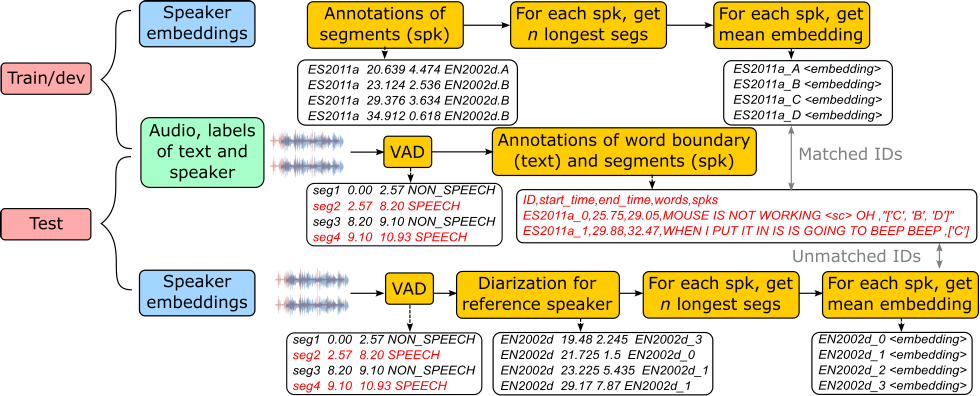}
  \caption{Preparation of the AMI corpus for the training, development, and test sets.}
  \label{fig:preparation}
\end{figure*}

\subsection{Overall system}
\label{subsec:system}

Our proposed system consists of VAD, SD, and SA-ASR modules. Figure~\ref{fig:system} illustrates the overall pipeline, which starts with VAD to divide the entire meeting into speech and non-speech segments. The VAD system takes as input the single-channel signal obtained by delay-and-sum (DAS) beamforming\textcolor{myred}{\cite{don1993array, speechbrain}} applied to the AMI 8-channel Multiple Distant Microphone (MDM) signal. Since SA-ASR requires speaker profiles as inputs, as mentioned in Section~\ref{subsec:saasr}, we need an SD model to determine the number of speakers in the entire meeting and identify the speech segments for each speaker. We use the results of VAD to perform SD on speech segments. Subsequently, based on the SD results, we use the non-overlapped portions of speech from each speaker to compute an average speaker embedding for that speaker. Finally, the segments obtained by VAD and the speaker profiles obtained through SD serve as input to SA-ASR, allowing us to retrieve the speech content of all different speakers in a first-in-first-out order. The SA-ASR system is single-channel and operates on the first channel of the AMI MDM signal.

\subsection{Data preparation}
\label{subsec:processing}
In real-life applications, ground-truth silence positions and utterance/word boundaries are not available at test time, so testing requires the use of VAD for segmentation. In \cite{cui2023}, the SA-ASR model fine-tuned on fixed-sized segments and tested on VAD segments was found to yield poor results, which can be attributed to mismatched fine-tuning and test segment lengths. To adapt the model to the length of VAD segments during test, we also utilize VAD segments during the training phase. The following outlines the preparation of training, development, and test data on the AMI corpus.

\subsubsection{Preparation of training, development, and test sets}
\label{subsubsec:preparation}
The process of obtaining speech segments, speaker, and text labels is the same for training, development, and test data. As shown in Figure~\ref{fig:preparation}, all meeting data is segmented into variable-sized speech segments using VAD. We merge adjacent speech segments separated by a silence shorter than a given duration threshold, resulting in an average segment length which depends on the specified threshold. Using AMI's annotation files with information about speaker segments and word boundaries, we assign text and speaker sequences to each segment in a first-in-first-out order.

The process of obtaining speaker profiles differs for training and development vs.\ test. For the training and validation sets, we extract all speech segments for each speaker in each meeting based on the ground-truth annotated speaker segments, and select the $n$ longest segments for each speaker. We use a pretrained speaker embedding model to embed all $n$ segments and calculate the average embedding as the template for the associated speaker. The process for the test set is different because, in real test scenarios, the number of speakers is not known a priori. To solve this issue, we apply the SD model to the VAD segments to estimate the number of speakers and the speech segments for each speaker. The speaker profiles for all speakers are then derived in the same way as for the training and validation sets.

\subsubsection{Remapping of speaker IDs on the test set}
\label{subsubsec:remapping}

When computing speaker profiles for the training and development sets, the speaker IDs align with the labels used for speaker identification, as they both rely on the same speaker segment annotation file. 
\textcolor{myred}{However, during test, the SD results may predict a different number of speakers and assign new IDs to the recognized speakers. 
This necessitates remapping SD speaker IDs to ground-truth speaker IDs for evaluation purposes.
This remapping process involves using the Hungarian algorithm \cite{kuhn1955hungarian}, as implemented by \cite{speechbrain}, during the calculation of the diarization error rate.\footnote{\url{https://github.com/speechbrain/speechbrain/blob/develop/tools/der_eval/md-eval.pl}}  } 

\section{Experimental settings}
\label{sec:exp}
\subsection{Dataset and metrics}
\label{subsec:data}
\subsubsection{Simulated multi-speaker LibriSpeech corpus}
\label{subsubsec:librispeech}
To achieve optimal performance in AMI, it is imperative to undergo pretraining with a larger and cleaner dataset \cite{kanda2021large,yang2023simulating,cui2023}.
We pretrained the SA-ASR model on distant-microphone multi-speaker data simulated using the LibriSpeech corpus \cite{panayotov2015LibriSpeech}. 
We utilized the train-960 and dev-clean subsets to construct our training and development datasets, creating a far-field microphone and multi-speaker set with 960 hours for training and 20 hours for development.
We generate Room impulse responses (RIRs) using the gpuRIR toolkit \cite{diaz2021gpurir}. The length, width and the height of the room are randomly drawn in the range of 3 to 8~m and 2.4 to 3~m, respectively. The microphone and speaker positions are randomly sampled with the constraints that the microphone is at most 0.5~m away from the room center, the speakers are at least 0.5~m away from the walls, and their heights are 0.6 to 0.8~m above the middle plane. The RT60 value is randomly drawn in the range of 0.4 to 1~s. \textcolor{myred}{Each} multi-speaker signal is generated by randomly drawing one utterance from each of 1 to 3 speakers, mixing these utterances convolved by RIRs, and concatenating the corresponding transcripts separated by the  \textless sc\textgreater\ token. The start time of each utterance is shifted relative to the previous one, with a delay ranging from 0.5 seconds to the maximum duration of the previous utterance.
This realistic choice also guarantees the first-in-first-out principle behind SOT. 
Each multi-speaker signal is linked to 8 speaker profiles\footnote{\textcolor{myred}{We chose a higher number to train a robust model, as real-life diarization results may identify more speakers than actually present.}}, encompassing both the actual speakers in the signal and templates from randomly selected speakers out of the 2,484 LibriSpeech speakers. Speaker embeddings are computed as the average of two random enrollment sentences for each speaker.

\subsubsection{AMI meeting corpus}
\label{subsubsec:ami}
We conducted fine-tuning and testing on the AMI corpus, which consists of approximately 100~h of multiple distant microphone (MDM) recordings with 3 to 5 speakers. In this paper, the VAD module takes \textcolor{myred}{the DAS beamformed signal obtained from} 8 MDM channels as input, while the SA-ASR module takes only the first MDM channel. We assessed three alternative segmentation methods. The first method, following \cite{cui2023}, involves fixed-sized segmentation with a chunk size of 5~s and a 2~s hop size, followed by adjusting the start/end times to non-overlapped word boundaries as described in Section \ref{subsubsec:fixed}. The second method, following \cite{kanda2021large, yang2023simulating}, involves segmenting based on ground-truth silence/non-speech positions between utterances: we extracted segments for all speakers from the ground-truth speaker and utterance boundary annotations, merged overlapped segments, and discarded the resulting segments which were longer than 100~s.

The third segmentation method described in Section~\ref{subsec:processing} uses VAD to extract the segments. 
We fused adjacent segments based on silence duration thresholds of 0.1, 0.3, 0.5, 0.7, and 0.9~s, resulting in datasets with different numbers of segments and different average segment lengths. Table~\ref{table:joint} provides the statistics and average lengths for the training set under all segmentation methods.

\subsubsection{Metrics}
\label{subsubsec:metrics}
We evaluate the SA-ASR system's ASR and speaker assignment performance using the word error rate (WER) and the token-level speaker error rate (SER)\footnote{The token-level SER is calculated from sequences of estimated and ground-truth token-level speaker labels, in the same way as the WER.}, respectively. Furthermore, we compute the speaker counting accuracy \cite{kanda2020joint,cui2023} \textcolor{myred}{in terms of confusion scores calculated as} 
\begin{align}\label{eq:eq2}
\textcolor{myred}{\text{Confusion\_score} (i, k)} = \frac{N^i_{k}}{N_k},
\end{align}
where $N^i_{k}$ is the number of test signals where the estimate indicates $i$ speakers when the ground truth has $k$ speakers, and ${N_k}$ represents the number of test signals where the ground truth has $k$ speakers.

\subsection{Model and training setup}
\label{subsec:model}
Our experiments were implemented using the SpeechBrain toolkit \cite{speechbrain}. For VAD, we used the CRDNN model\footnote{Available at \url{https://huggingface.co/speechbrain/vad-crdnn-libriparty}} pretrained on Libriparty \cite{LibriParty}. For SD, we use x-vectors as speaker embeddings and the Spectral Clustering \cite{ning2006spectral} method. The SD and SA-ASR systems share a common speaker embedding model, ECAPA-TDNN\footnote{Available at \url{https://huggingface.co/speechbrain/spkrec-ecapa-voxceleb}}, which is pretrained on the VoxCeleb1 \cite{chung2019voxsrc} and VoxCeleb2 \cite{nagrani2020voxsrc} training datasets, yielding 192-dimensional embeddings. Our text tokenizer is a SentencePiece model \cite{kudo2018sentencepiece} with a vocabulary of 5,000 tokens. In SA-ASR, the Conformer-based encoder, the Transformer-based decoder, and the speaker decoder have 12, 6, and 2 layers, respectively. All multi-head attention mechanisms have 4 heads, the model dimension is set to 256, and the size of the feedforward layer is 2,048. The number of parameters of the employed CRDNN, ECAPA-TDNN, and SA-ASR models is 0.1M, 22.15M and 61.1M, respectively.

The ASR module in SA-ASR was pretrained on simulated multi-speaker Librispeech data for 360k iterations using the Adam optimizer with a learning rate of $5\times 10^{-4}$. The ASR and speaker modules in SA-ASR were then further pretrained for 300k iterations using a learning rate of $2.5\times 10^{-4}$. Finally, we fine-tuned the model on real AMI single distant microphone (SDM) data for 20k iterations, using the Adam optimizer with a learning rate of $3\times 10^{-4}$. We tuned only the ASR module for the first 10k iterations and performed joint tuning of the ASR and speaker modules in the last 15k iterations. Those numbers of iterations were fixed in preliminary experiments based on the development set WER.

\begin{table*}[t]
\caption{WER and SER (\%) on the AMI test set using different segmentation methods for SA-ASR fine-tuning and testing. The speaker profiles are computed by averaging the embeddings obtained from the 3 longest annotated segments.}
\label{table:joint}
\centering
\scalebox{0.9}{
\addtolength{\tabcolsep}{-0.5em}
\begin{tabular}{cccccccccccccccccc}
    \toprule
    \multirow{3.8}{*}{\begin{minipage}{2cm}\centering\bfseries Train/dev\\seg method\end{minipage}} & 
    \multicolumn{3}{c}{\bfseries Train set statistics} &
    \multicolumn{4}{c}{\bfseries Test (matched seg method)} &
    \multicolumn{10}{c}{\bfseries Test (VAD seg with 0.5~s silence threshold)} 
    \\ 
    \cmidrule(lr){5-8} \cmidrule(lr){9-18}
    &
    &&&\multicolumn{4}{c}{\bfseries 1,2,3,4-spk mix }&
    \multicolumn{2}{c}{\bfseries 1-spk } & 
    \multicolumn{2}{c}{\bfseries 2-spk mix } & 
    \multicolumn{2}{c}{\bfseries 3-spk mix }& 
    \multicolumn{4}{c}{\bfseries 1,2,3,4-spk mix}
    \\ 
    \cmidrule(lr){2-4} \cmidrule(lr){5-8}   \cmidrule(lr){9-10} \cmidrule(lr){11-12} \cmidrule(lr){13-14}  \cmidrule(lr){15-18}
    &\textbf{Sil.\ thresh}& \textbf{\# Seg}& \textbf{Avg dur}& \textbf{\# Seg}& \textbf{Avg dur}& \textbf{WER}& \textbf{SER}& \textbf{WER}& \textbf{SER}& \textbf{WER}& \textbf{SER}& \textbf{WER}& \textbf{SER}& \textbf{\# Seg}& \textbf{Avg dur}& \textbf{WER}& \textbf{SER}\\ 
    \cmidrule(lr){1-18}
    \multirow{1}{*}{Fixed-size (5~s)} & - &90,426&6.59~s&10,234&6.63~s&44.54&28.54&49.41&28.92&50.91&29.78&57.48&36.29&\multirow{1}{*}{3,586} &\multirow{1}{*}{6.96~s} &55.91&34.94\\
    \hline
   Ground-truth&-&21,917&10.19~s&2,504&8.67~s&\textbf{43.42}&\textbf{21.81}&44.47&23.08&50.74&28.21&56.77&31.47&3,586&6.96~s&54.00&31.02\\
   \hline
   \multirow{5}{*} {VAD}& 0.1~s & 93,150 &1.59~s&10,605&1.59~s& 44.62&29.03&\textbf{30.53}&17.12&45.67&29.15&53.04&38.07&\multirow{5}{*}{3,586} &\multirow{5}{*}{6.96~s}&47.02&30.19\\
    &0.3~s & 57,589 &3.40~s &6,095&3.71~s& 44.50&28.85& 31.95& 16.17& 43.54 & 28.99& 52.38 & 35.69& &&46.33&28.99\\
    &0.5~s & 35,091 &6.18~s &3,586&6.96~s&45.09& 27.46&31.25&15.53&42.05&23.21&50.57&30.54&&&\textbf{45.09}&27.46\\
    &0.7~s & 22,027 &10.33~s&2,265&10.24~s&45.80&24.82&32.13&\textbf{13.51}&\textbf{41.74}&\textbf{22.17}&\textbf{49.82}&\textbf{27.00}&&&46.13&27.02 \\
    &0.9~s & 14,460&15.65~s &1,519&13.15~s&47.08&24.95&33.05&\textbf{13.35}&42.52&\textbf{22.31}&51.39&\textbf{27.30}&&&46.27&\textbf{25.04} \\
    \bottomrule
\end{tabular}
}
\begin{tablenotes}
\centering
      \footnotesize
      \item \textcolor{black}{Note: For all the tables, we employed the SCTK toolkit \cite{sctk} to conduct significance tests, specifically the Matched Pair Sentence Segment test. We highlight in bold the best WER/SER result and the results statistically equivalent to it at a 0.05 significance level.}
    \end{tablenotes}
\end{table*}

\section{Discussion of results}
\label{sec:discuss}

\subsection{Effectiveness of fine-tuning on VAD segments}
\label{subsec:effectiveness}

In practical applications, it is necessary to divide long continuous meeting audio into smaller-sized segments. As detailed in Section~\ref{subsubsec:ami}, our experiments involve three alternative segmentation methods. When fine-tuning the SA-ASR system, we employed different training data segmented using these three distinct methods. To emulate real-life test conditions, the primary focus is on testing all models using VAD output segments. The results presented in Table~\ref{table:joint} showcase the results on VAD segments with 0.5~s silence threshold, consistent with the subsequent tables. To gain a more comprehensive understanding of the model's capabilities across various speaker counts, we report the results on subsets comprising 1 to 3 speakers. For comparison, we also report the performance of each model on a matched test segmented in the same way as the training set.

Looking at the last column in Table~\ref{table:joint}, models fine-tuned on VAD segments consistently exhibit superior performance when tested on VAD segments compared to models fine-tuned on fixed-sized or ground-truth segments. The relative WER reduction can be up to 19\% (from 55.91\% to 45.09\%), and the relative SER reduction can be as high as 28\% (from 34.94\% to 25.04\%). However, this difference is not as pronounced when models are tested on their respective matched test sets. This underscores the significance of training on data that closely aligns with the test data type. For real-life applications relying on VAD segments, it is advisable to fine-tune on VAD segments, even when speaker segment annotations are available in the training set. It is noteworthy to highlight that, even with comparable average segment lengths (10.19~s and 10.33~s), fine-tuning on VAD segments can result in a 15\% (from 54\% to 46.13\%) lower relative WER and a 13\% (from 31.02\% to 27.02\%) lower relative SER compared to the model fine-tuned on ground-truth segments. 



\begin{table}[!t]
\caption{Speaker counting accuracy (\%) on the AMI test set with 0.5~s VAD silence threshold. \textcolor{myred} { We mapped the speaker number from SD output to have an evaluation without SA-ASR. } }
%
\label{table:spks-count-ami}
\centering
\scalebox{0.9}{
\addtolength{\tabcolsep}{-0.5em}
\begin{tabular}{c|cccccccc}
\toprule
\multirow{2}{*}{\bfseries } &
    \multirow{2.4}{*}{\bfseries Method} & 
    \multirow{2}{*}{\bfseries \# Speakers}& 
    \multicolumn{5}{c}{\bfseries Estimated \# speakers } & 
    \\ 
    \cmidrule(lr){4-8}
    &&&  1&2&3&4&\textgreater4
    \\ 
    \cmidrule(lr){1-8}
     \multirow{4}{*} {\rotatebox[origin=c]{90}{\begin{minipage}{1.4cm}\centering\bfseries Without \\SA-ASR\end{minipage}}}&\multirow{4}{*}{\textcolor{myred}{SD}} & 1 & 89.20& 10.27& 0.53& 0.00 &0.00\\
  &  & 2  &59.18 & 35.37 & 5.34& 0.11 &0.00 \\
  &   & 3  & 29.73 & 48.65& 19.46&2.16&0.00\\
   &   & 4  & 10.24 &39.37& 33.07 & 14.57&2.76\\
    \hline
    \multirow{12}{*} {\rotatebox[origin=c]{90}{\begin{minipage}{4cm}\centering\bfseries SA-ASR \\Train/dev seg method\end{minipage}}}
&
     \multirow{4}{*} {Fixed-size (5~s)} & 1 & 80.04 & 17.28& 2.29& 0.32 &0.05\\
  &   & 2  & 14.09 & \textbf{69.01}& 16.28 & 0.54 &0.00 \\
  &   & 3  & 0.73& 38.93&\textbf{50.04}&9.18&0.91 \\
 &     & 4  & 0.40 &15.98& 56.96 & 20.90&4.50\\
     \cline{2-8}
    &
    \multirow{4}{*} {Ground-truth} & 1 & 89.00 & 9.66& 1.22& 0.10 &0.00\\
   &  & 2  & 20.41 & 63.53& 15.28& 0.54 &0.10\\
    & & 3  &2.54 & 35.39& 48.63&12.15&0.90 \\
    &  & 4  & 0.00 &11.47& 54.50 & \textbf{29.50}&3.27\\
     \cline{2-8}
    & & 1 & \textbf{94.76 }& 4.91& 0.26& 0.05 &0.00\\
  &  VAD & 2  & 21.24 & 67.68 & 10.07 & 0.87 &0.00 \\
   & (0.5~s sil.\ thresh) & 3  & 2.72 & 40.29& 47.00&8.89&0.72\\
    &  & 4  & 0.00 &11.57& 59.91 & 23.55&3.71\\
     
\bottomrule
\end{tabular}
}
\end{table}

Focusing on the models fine-tuned on VAD segments in Table~\ref{table:joint}, it can be seen that the model fine-tuned on segments with 0.5~s VAD silence threshold exhibits a lower WER, which can be attributed to the matched fine-tuning and test segment lengths. However, the SER shows a different trend. Specifically, as the average fine-tuning segment length becomes longer, the corresponding SER decreases. The SER exhibits a relative reduction of 22\% (from 17.12\% to 13.35\%), 23\% (from 29.15\% to 22.31\%), and 28\% (from 38.07\% to 27.30\%) for 1, 2, and 3 speakers, respectively, when transitioning from an average fine-tuning length of 1.59~s to 15.65~s. This indicates that, as the training segments become longer, SA-ASR demonstrates improved capability in assigning speaker identities. 

Table~\ref{table:spks-count-ami} reports the speaker counting accuracy of the three systems. \textcolor{myred} {First of all, the assignment of speaker labels solely based on SD is not accurate, particularly for multi-speaker scenarios. This underscores the critical role of SA-ASR systems. For the comparison of SA-ASR models,} on the 1-speaker test subset, the model fine-tuned on VAD segments exhibits the highest accuracy at 94.76\%, compared 80.01\% for the model fine-tuned on fixed-sized segments. However, the latter system demonstrates slightly higher accuracy when the number of speakers is 2  (69.01\%) or 3 (50.04\%). The model fine-tuned on ground-truth segments performs better when the number of speakers is as high as 4 (29.50\%). Future studies can explore how to employ different training techniques for datasets with specific numbers of speakers.

\begin{table}[!b]
\caption{\textcolor{myred} {SER (\%) on the AMI test set with 0.5~s VAD silence threshold as a function of the speaker profiles used for fine-tuning and testing. Ref: templates extracted from annotated speaker segments. Test: templates extracted from SD output. The WER is unaffected and equal to that in Table~\ref{table:joint} (1-spk 31.25\%, 2-spk mix 42.05\%, 3-spk mix 50.57\%, total 45.09\%).}}
\label{table:diarization}
\centering
\addtolength{\tabcolsep}{-0.15em}

\begin{tabular}{ccccccccccc}
    \toprule
    {\bfseries Train/dev} &   
    {\bfseries Test} & 
    {\bfseries 1-spk} & 
    {\bfseries 2-spk mix} & 
    {\bfseries 3-spk mix}& 
    {\bfseries Total}
    \\ 
    \cmidrule(lr){1-6}
    Ref & Ref &15.53&23.21&30.54&27.46\\
    Ref & Test &\textbf{12.97}&\textbf{20.76}&\textbf{27.43}&\textbf{26.54}\\

    \bottomrule
\end{tabular}
\end{table}

\begin{table*}[t]
\caption{\textcolor{myred} {SER (\%) on the AMI test set with 0.5~s VAD silence threshold as a function of the VAD silence threshold and the length of candidate segments used to extract non-overlapped speaker profiles. The WER is unaffected and equal to that in Table~\ref{table:joint}.}}
\label{table:vad-ser}
\centering
\scalebox{0.9}{
\addtolength{\tabcolsep}{-0.2em}
\begin{tabular}{cc|ccc|ccc|cccccc}
    \toprule
    \multicolumn{2}{c}{\bfseries VAD segments} &
    \multicolumn{3}{c}{\bfseries Using all 2--5~s segments} & 
    \multicolumn{3}{c}{\bfseries Using all 5--10~s segments} &
    \multicolumn{3}{c}{\bfseries Using all 6--50~s segments} 
    \\ 
      \cmidrule(lr){1-2}\cmidrule(lr){3-5}\cmidrule(lr){6-8}\cmidrule(lr){9-11}
   \textbf{Sil.\ thresh} & \textbf{Avg dur} &\textbf{1-spk}& \textbf{2-spk mix}& \textbf{3-spk mix}&\textbf{1-spk}& \textbf{2-spk mix}& \textbf{3-spk mix}&
    \textbf{1-spk}& \textbf{2-spk mix}& \textbf{3-spk mix}\\
    \cmidrule(lr){1-11}
    0.1~s &1.59~s&\textbf{12.79}&\textbf{20.62}&\textbf{27.44}&60.53 \ding{61}&54.96 \ding{61}&\ding{61}&\ding{61}&\ding{61}&\ding{61}\\
    0.3~s &3.40~s&13.09&20.85&\textbf{27.04}&14.16&21.76&28.08&19.03 \ding{61}&26.18 \ding{61}&32.48 \ding{61}\\
     0.5~s &6.18~s&\textbf{12.83}&20.93 &27.59&13.88&21.82&27.97&\textbf{12.97}&\textbf{20.76}&\textbf{27.43}\\
    0.7~s &10.33~s&14.26&22.46&27.71&14.17&22.53&27.83&\textbf{12.76}&20.98&27.71\\
    0.9~s &15.65~s&14.58&22.16&27.84&\textbf{12.94}&21.53&\textbf{27.49}&14.11&21.75&27.81\\
    \bottomrule
\end{tabular}
}
\begin{tablenotes}
\centering
      \small
      \item \ding{61}: The VAD output segments with 0.1~s or 0.3~s silence threshold are inadequate for obtaining enough candidate segments with the specified length, resulting in unavailable or abnormal speaker assignment results.
    \end{tablenotes}
\end{table*}

\subsection{Extraction of speaker profiles from SD output}
\label{subsec:accuracy}
So far, we have used annotated speaker segments to extract the speaker profiles in the training, development, and test sets. In a real-life situation, using SD segments to extract the speaker profiles in the test set is desirable. However, this leads to statistical differences between the fine-tuning and test phases. To understand how this affects the SER, we evaluate both types of templates in the test set. The resulting performance when fine-tuning and testing on VAD output segments with 0.5~s silence threshold is presented in Table~\ref{table:diarization}.

Surprisingly, while the model is fine-tuned on speaker profiles extracted from annotated speaker segments, extracting speaker profiles from SD segments during test reduces the SER by up to \textcolor{myred} {16\% relative (from 15.53\% to 12.97\%)}. We attribute this to the fact that human segmentation is partly inaccurate, especially when marking the boundaries of each speaker's speech segments. By contrast, SD can offer more precise speech boundaries, especially for the longest non-overlapped segments which are utilized to compute the speaker profiles.

Surprisingly too, these improved speaker profiles do not affect the WER, despite the fact that the derived token-level weighted speaker profiles are improved and fed back to the ASR Decoder (see Figure~\ref{fig:system}). This suggests that this feedback mechanism proposed in \cite{kanda2021end} may not fulfil its goal of enabling the Speaker Decoder to assist the ASR Decoder.

\subsection{Extraction of speaker profiles from different segment lengths}
\label{subsec:extraction}
Finally, we explored the extraction of speaker profiles from varying segment lengths and with different numbers of candidate segments. To begin with, the impact of using VAD with 0.1~s to 0.9~s silence threshold as input segments to the SD system is analyzed in Table~\ref{table:vad-ser}. For each SD result, we filtered candidate segments for each speaker by considering different length intervals. The mean embedding for each speaker was then calculated from all segments in that length interval. The results showcased in Table~\ref{table:vad-ser} reveal several good solutions with either short or long candidate segment lengths. When candidate segment lengths fall within the range of 2 to 5~s, VAD silence thresholds below 0.5~s are preferable to longer ones, with a relative SER reduction of up to \textcolor{myred} {12\% (12.79\% compared to 14.58\%)}. On the contrary, when using longer candidate segments ranging from 6 to 50~s, short VAD silence thresholds are inadequate and VAD silence thresholds above 0.5~s are preferable.

\begin{figure}[!ht]
  \centering
  \includegraphics[width=1\linewidth]{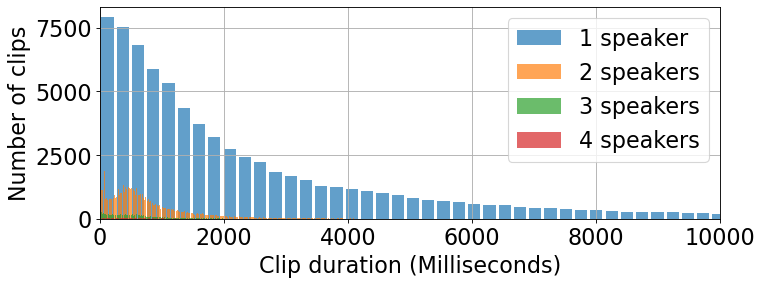}
  \caption{Histograms of speaker overlaps on the AMI corpus. Only clips shorter than 10~s are shown.}
  \label{fig:spk-overlap}
\end{figure}

We further examined how different configurations, including the presence of overlapping segments, the number of candidate segments, and the length of candidate segments, impact the speaker profiles and the resulting SER. Table~\ref{table:numspk} illustrates the influence of these attributes using VAD segments with 0.5~s silence threshold as inputs to SD. Using all candidate segments instead of the 2 or 10 longest leads to an enhanced speaker representation irrespective of the segment length, as evidenced by a lower SER up to \textcolor{myred} {15\% relative (from 15.02\% to 12.83\%)}. Moreover, shorter candidate segments exhibit a greater sensitivity to the number of segments. This can be attributed to the fact that a larger number of segments enables a more robust and comprehensive representation of each speaker. The presence of overlapping segments among the candidate segments does not appear to have a significant impact. 
Using segment annotations, we compute the number of overlapping clips involving varying numbers of speakers. Subsequently, for each speaker count, we calculate the quantity of overlapping clips across different durations of overlap. Figure~\ref{fig:spk-overlap} presents statistical findings across the complete AMI corpus, revealing a minimal occurrence of 2, 3, and 4 overlapping speaker regions.

\begin{table}[!t]
\caption{\textcolor{myred} {SER (\%) on the AMI test set with 0.5 s VAD silence threshold as a function of the number and the length of candidate segments used to extract speaker profiles. The WER is unaffected and equal to that in Table~\ref{table:joint}.}}
\label{table:numspk}
\centering
\scalebox{0.9}{
\addtolength{\tabcolsep}{-0.2em}
\begin{tabular}{ccc|ccccccccc}
    \toprule
    \multicolumn{3}{c}{\bfseries Candidate segments} & 
    
    \multicolumn{3}{c}{\bfseries SER} & 
    \\ 
     \cmidrule(lr){1-3}\cmidrule(lr){4-6}
    \textbf{Overlap}&\textbf{\# seg}&\textbf{Dur }&
    \textbf{1-spk}& \textbf{2-spk mix}& \textbf{3-spk mix}\\
    \cmidrule(lr){1-6}


   no& 2 & 2--5~s  &15.02&23.72 &29.75\\
   no& 10 & 2--5~s &14.44&21.92 &27.82 \\
   no& all & 2--5~s &\textbf{12.83}&\textbf{20.93} &27.59 \\
   yes& all & 2--5~s &\textbf{12.80}&\textbf{20.81}&\textbf{27.28}  \\
   \hline
   no&2 & 5--10~s  &13.09&21.29 &28.00 \\
    no&10 & 5--10~s  &14.03&21.69&27.76 \\
    no&all & 5--10~s  &13.88&21.82&27.97 \\
    yes&all & 5--10~s  &13.93&21.72&27.92 \\
    \hline
         no&2 & 6--50~s  &13.04&21.42&27.53\\
     no&10 & 6--50~s  &13.33&21.08&\textbf{27.13}\\
    no&all & 6--50~s  &\textbf{12.97}&\textbf{20.76}&\textbf{27.43}\\
       yes&all & 6--50~s  &13.06&\textbf{20.81}&\textbf{27.41}\\
    \bottomrule
\end{tabular}
}
\end{table}

\begin{figure}[!ht]
  \centering
  \includegraphics[width=1\linewidth]{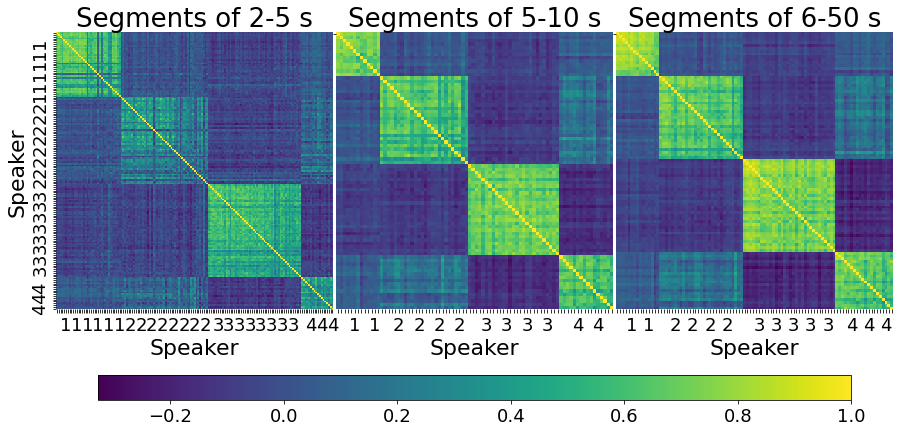}
  \caption{\textcolor{myred} {Similarity matrices between the speaker embeddings of the candidate segments in meeting ES2004c. With the three specified segment lengths, the number of candidate segments is 154, 81, and 96, and the resulting SER for that meeting is 19.88\%, 19.65\%, and 19.62\%, respectively.}}
  \label{fig:similarity}
\end{figure}

Figure~\ref{fig:similarity} depicts the cosine similarity between the speaker embeddings of the candidate segments in one meeting for the three specified segment lengths. Segments in the range of 6 to 50~s exhibit a higher similarity compared to those in the range of 2 to 5~s, leading to a slightly lower SER \textcolor{myred} {(19.62\% vs.\ 19.88\%)}.

Finally, note that here again the various speaker profiles explored in Tables~\ref{table:vad-ser} and \ref{table:numspk} do not affect the WER.


\section{Conclusion}
\label{sec:conclusion}
We focused on enhancing speaker assignment in SA-ASR for real-life meetings. We proposed a VAD-SD-SA-ASR pipeline and advocated for fine-tuning on VAD segments instead of fixed-size segments. We showed that this can lead to relative SER reduction up to 28\%. We then explored strategies for extracting speaker profiles by varying the VAD and SD configurations. Our results reveal that extracting them from SD output rather than annotated speaker segments results reduces the SER by up to \textcolor{myred} {16\% }relative. Moreover, while extracting speaker profiles from short or long segments can yield a similar SER, the latter results in improved speaker representation and speaker similarity across segments and is less sensitive to the number of segments averaged to obtain the template. Finally, we have found that improved speaker profiles do not affect the WER; this suggests that an effective feedback mechanism enabling the Speaker Decoder to assist the ASR Decoder is still to be found.


\section{Acknowledgements}
\label{sec:ack}
Experiments presented in this paper were carried out using (a) the Grid'5000 testbed, supported by a scientific interest group hosted by Inria and including CNRS, RENATER and several Universities as well as other organizations (see https://www.grid5000.fr), and (b) HPC resources from GENCI-IDRIS (Grant 2023-[AD011013881]).

\bibliographystyle{IEEEbib_abbrev}
\bibliography{Odyssey2024_BibEntries}

\begin{thebibliography}{10}

\bibitem{yu2022m2met}
F. Yu, S. Zhang, Y. Fu, L. Xie, S. Zheng, Z. Du, W. Huang, P. Guo, Z. Yan, B.
  Ma, et~al.,
\newblock ``M2met: The {ICASSP} 2022 multi-channel multi-party meeting
  transcription challenge,''
\newblock in {\em 2022 IEEE International Conference on Acoustics, Speech and
  Signal Processing (ICASSP)}, 2022, pp. 6167--6171.

\bibitem{yoshioka2019advances}
T. Yoshioka, I. Abramovski, C. Aksoylar, Z. Chen, M. David, D. Dimitriadis, Y.
  Gong, I. Gurvich, X. Huang, Y. Huang, et~al.,
\newblock ``Advances in online audio-visual meeting transcription,''
\newblock in {\em 2019 IEEE Automatic Speech Recognition and Understanding
  Workshop (ASRU)}, 2019, pp. 276--283.

\bibitem{yoshioka2019meeting}
T. Yoshioka, D. Dimitriadis, A. Stolcke, W. Hinthorn, Z. Chen, M. Zeng, and X.
  Huang,
\newblock ``Meeting transcription using asynchronous distant microphones.,''
\newblock in {\em Interspeech}, 2019, pp. 2968--2972.

\bibitem{cornell2023chime}
S. Cornell, M.~S. Wiesner, S. Watanabe, D. Raj, X. Chang, P. Garcia, Y.
  Masuyam, Z.-Q. Wang, S. Squartini, and S. Khudanpur,
\newblock ``The {CHiME-7 DASR Challenge}: Distant meeting transcription with
  multiple devices in diverse scenarios,''
\newblock in {\em 7th International Workshop on Speech Processing in Everyday
  Environments (CHiME)}, 2023, pp. 1--6.

\bibitem{9054029}
X. Chang, W. Zhang, Y. Qian, J. Le~Roux, and S. Watanabe,
\newblock ``End-to-end multi-speaker speech recognition with {Transformer},''
\newblock in {\em 2020 IEEE International Conference on Acoustics, Speech and
  Signal Processing (ICASSP)}, 2020, pp. 6134--6138.

\bibitem{seki-etal-2018-purely}
H. Seki, T. Hori, S. Watanabe, J. Le~Roux, and J.~R. Hershey,
\newblock ``A purely end-to-end system for multi-speaker speech recognition,''
\newblock in {\em 56th Annual Meeting of the ACL (Volume 1: Long Papers)},
  2018, pp. 2620--2630.

\bibitem{kanda2021end}
N. Kanda, G. Ye, Y. Gaur, X. Wang, Z. Meng, Z. Chen, and T. Yoshioka,
\newblock ``End-to-end speaker-attributed {ASR} with {Transformer},''
\newblock in {\em Interspeech}, 2021, pp. 4413--4417.

\bibitem{yu2022comparative}
F. Yu, Z. Du, S. Zhang, Y. Lin, and L. Xie,
\newblock ``A comparative study on speaker-attributed automatic speech
  recognition in multi-party meetings,''
\newblock in {\em Interspeech}, 2022, pp. 560--564.

\bibitem{chang2021hypothesis}
X. Chang, N. Kanda, Y. Gaur, X. Wang, Z. Meng, and T. Yoshioka,
\newblock ``Hypothesis stitcher for end-to-end speaker-attributed {ASR} on
  long-form multi-talker recordings,''
\newblock in {\em 2021 IEEE International Conference on Acoustics, Speech and
  Signal Processing (ICASSP)}, 2021, pp. 6763--6767.

\bibitem{kanda2021investigation}
N. Kanda, X. Chang, Y. Gaur, X. Wang, Z. Meng, Z. Chen, and T. Yoshioka,
\newblock ``Investigation of end-to-end speaker-attributed {ASR} for continuous
  multi-talker recordings,''
\newblock in {\em 2021 IEEE Spoken Language Technology Workshop (SLT)}, 2021,
  pp. 809--816.

\bibitem{kanda2022transcribe}
N. Kanda, X. Xiao, Y. Gaur, X. Wang, Z. Meng, Z. Chen, and T. Yoshioka,
\newblock ``Transcribe-to-diarize: Neural speaker diarization for unlimited
  number of speakers using end-to-end speaker-attributed {ASR},''
\newblock in {\em 2022 IEEE International Conference on Acoustics, Speech and
  Signal Processing (ICASSP)}, 2022, pp. 8082--8086.

\bibitem{kanda2022streaming}
N. Kanda, J. Wu, Y. Wu, X. Xiao, Z. Meng, X. Wang, Y. Gaur, Z. Chen, J. Li, and
  T. Yoshioka,
\newblock ``Streaming speaker-attributed {ASR} with token-level speaker
  embeddings,''
\newblock in {\em Interspeech}, 2022, pp. 521--525.

\bibitem{guo2021multi}
P. Guo, X. Chang, S. Watanabe, and L. Xie,
\newblock ``Multi-speaker {ASR} combining non-autoregressive {Conformer CTC}
  and conditional speaker chain,''
\newblock in {\em Interspeech}, 2021, pp. 3720--3724.

\bibitem{kanda2021large}
N. Kanda, G. Ye, Y. Wu, Y. Gaur, X. Wang, Z. Meng, Z. Chen, and T. Yoshioka,
\newblock ``Large-scale pre-training of end-to-end multi-talker {ASR} for
  meeting transcription with single distant microphone,''
\newblock in {\em Interspeech}, 2021, pp. 3430--3434.

\bibitem{cui2023}
C. Cui, I. Sheikh, M. Sadeghi, and E. Vincent,
\newblock ``End-to-end multichannel speaker-attributed {ASR}: Speaker guided
  decoder and input feature analysis,''
\newblock in {\em 2023 IEEE Automatic Speech Recognition and Understanding
  Workshop (ASRU)}, 2023, pp. 1--8.

\bibitem{yang2023simulating}
M. Yang, N. Kanda, X. Wang, J. Wu, S. Sivasankaran, Z. Chen, J. Li, and T.
  Yoshioka,
\newblock ``Simulating realistic speech overlaps improves multi-talker {ASR},''
\newblock in {\em 2023 IEEE International Conference on Acoustics, Speech and
  Signal Processing (ICASSP)}, 2023, pp. 1--5.

\bibitem{carletta2005ami}
J. Carletta, S. Ashby, S. Bourban, M. Flynn, M. Guillemot, T. Hain, J. Kadlec,
  V. Karaiskos, W. Kraaij, M. Kronenthal, et~al.,
\newblock ``The {AMI} meeting corpus: A pre-announcement,''
\newblock in {\em International Workshop on Machine Learning for Multimodal
  Interaction}, 2005, pp. 28--39.

\bibitem{medennikov2020target}
I. Medennikov, M. Korenevsky, T. Prisyach, Y. Khokhlov, M. Korenevskaya, I.
  Sorokin, T. Timofeeva, A. Mitrofanov, A. Andrusenko, I. Podluzhny, A. Laptev,
  and A. Romanenko,
\newblock ``Target-speaker voice activity detection: A novel approach for
  multi-speaker diarization in a dinner party scenario,''
\newblock in {\em Interspeech}, 2020, pp. 274--278.

\bibitem{yang2010comparative}
X. Yang, B. Tan, J. Ding, J. Zhang, and J. Gong,
\newblock ``Comparative study on voice activity detection algorithm,''
\newblock in {\em 2010 International Conference on Electrical and Control
  Engineering}, 2010, pp. 599--602.

\bibitem{park2022review}
T.~J. Park, N. Kanda, D. Dimitriadis, K.~J. Han, S. Watanabe, and S. Narayanan,
\newblock ``A review of speaker diarization: Recent advances with deep
  learning,''
\newblock {\em Computer Speech \& Language}, vol. 72, pp. 101317, 2022.

\bibitem{xiao2021microsoft}
X. Xiao, N. Kanda, Z. Chen, T. Zhou, T. Yoshioka, S. Chen, Y. Zhao, G. Liu, Y.
  Wu, J. Wu, et~al.,
\newblock ``Microsoft speaker diarization system for the {VoxCeleb} speaker
  recognition challenge 2020,''
\newblock in {\em 2021 IEEE International Conference on Acoustics, Speech and
  Signal Processing (ICASSP)}, 2021, pp. 5824--5828.

\bibitem{snyder2018x}
D. Snyder, D. Garcia-Romero, G. Sell, D. Povey, and S. Khudanpur,
\newblock ``X-vectors: Robust {DNN} embeddings for speaker recognition,''
\newblock in {\em 2018 IEEE International Conference on Acoustics, Speech and
  Signal Processing (ICASSP)}, 2018, pp. 5329--5333.

\bibitem{asif2022emotion}
M. Asif, M.~T. Vinodbhai, S. Mishra, A. Gupta, and U.~S. Tiwary,
\newblock ``Emotion recognition in {VAD} space during emotional events using
  {CNN-GRU} hybrid model on {EEG} signals,''
\newblock in {\em International Conference on Intelligent Human Computer
  Interaction}. Springer, 2022, pp. 75--84.

\bibitem{lin202347}
J. Lin, K.-F. Un, W.-H. Yu, R.~P. Martins, and P.-I. Mak,
\newblock ``A {47-nW }voice activity detector ({VAD}) featuring a short-time
  {CNN} feature extractor and an {RNN-Based} classifier with a non-volatile
  {CAP-ROM},''
\newblock {\em IEEE Journal of Solid-State Circuits}, vol. 58, pp. 3020--3029,
  2023.

\bibitem{kim2016vowel}
J. Kim, J. Kim, S. Lee, J. Park, and M. Hahn,
\newblock ``Vowel based voice activity detection with {LSTM} recurrent neural
  network,''
\newblock in {\em Proceedings of the 8th International Conference on Signal
  Processing Systems}, 2016, pp. 134--137.

\bibitem{sainath2015convolutional}
T.~N. Sainath, O. Vinyals, A. Senior, and H. Sak,
\newblock ``Convolutional, long short-term memory, fully connected deep neural
  networks,''
\newblock in {\em 2015 IEEE International Conference on Acoustics, Speech and
  Signal Processing (ICASSP)}, 2015, pp. 4580--4584.

\bibitem{xiang2021fast}
Y. Xiang, T. Tang, T. Su, C. Brach, L. Liu, S.~S. Mao, and M. Geimer,
\newblock ``{Fast CRDNN: Towards on site training of mobile construction
  machines},''
\newblock {\em IEEE Access}, vol. 9, pp. 124253--124267, 2021.

\bibitem{dehak2010front}
N. Dehak, P.~J. Kenny, R. Dehak, P. Dumouchel, and P. Ouellet,
\newblock ``Front-end factor analysis for speaker verification,''
\newblock {\em IEEE Transactions on Audio, Speech, and Language Processing},
  vol. 19, no. 4, pp. 788--798, 2010.

\bibitem{variani2014deep}
E. Variani, X. Lei, E. McDermott, I.~L. Moreno, and J. Gonzalez-Dominguez,
\newblock ``Deep neural networks for small footprint text-dependent speaker
  verification,''
\newblock in {\em 2014 IEEE International Conference on Acoustics, Speech and
  Signal Processing (ICASSP)}, 2014, pp. 4052--4056.

\bibitem{desplanques2020ecapa}
B. Desplanques, J. Thienpondt, and K. Demuynck,
\newblock ``{ECAPA-TDNN}: Emphasized channel attention, propagation and
  aggregation in {TDNN} based speaker verification,''
\newblock in {\em Interspeech}, 2020, pp. 3830--3834.

\bibitem{ning2006spectral}
H. Ning, M. Liu, H. Tang, and T.~S. Huang,
\newblock ``A spectral clustering approach to speaker diarization,''
\newblock in {\em Interspeech}, 2006, pp. 2178--2181.

\bibitem{kanda2020serialized}
N. Kanda, Y. Gaur, X. Wang, Z. Meng, and T. Yoshioka,
\newblock ``Serialized output training for end-to-end overlapped speech
  recognition,''
\newblock in {\em Interspeech}, 2020, pp. 2797--2801.

\bibitem{don1993array}
D.~H. Johnson and D.~E. Dudgeon,
\newblock {\em Array signal processing: concepts and techniques},
\newblock Prentice Hall., 1993.

\bibitem{speechbrain}
M. Ravanelli, T. Parcollet, P. Plantinga, A. Rouhe, S. Cornell, L. Lugosch, C.
  Subakan, N. Dawalatabad, A. Heba, J. Zhong, J.-C. Chou, S.-L. Yeh, S.-W. Fu,
  C.-F. Liao, E. Rastorgueva, F. Grondin, W. Aris, H. Na, Y. Gao, R.~D. Mori,
  and Y. Bengio,
\newblock ``{SpeechBrain}: A general-purpose speech toolkit,'' 2021,
\newblock arXiv:2106.04624.

\bibitem{kuhn1955hungarian}
H.~W. Kuhn,
\newblock ``The hungarian method for the assignment problem,''
\newblock {\em Naval research logistics quarterly}, vol. 2, no. 1-2, pp.
  83--97, 1955.

\bibitem{panayotov2015LibriSpeech}
V. Panayotov, G. Chen, D. Povey, and S. Khudanpur,
\newblock ``Librispeech: An {ASR} corpus based on public domain audio books,''
\newblock in {\em 2015 IEEE International Conference on Acoustics, Speech and
  Signal Processing (ICASSP)}, 2015, pp. 5206--5210.

\bibitem{diaz2021gpurir}
D. Diaz-Guerra, A. Miguel, and J.~R. Beltran,
\newblock ``{gpuRIR}: A {Python} library for room impulse response simulation
  with {GPU} acceleration,''
\newblock {\em Multimedia Tools and Applications}, vol. 80, pp. 5653--5671,
  2021.

\bibitem{kanda2020joint}
N. Kanda, Y. Gaur, X. Wang, Z. Meng, Z. Chen, T. Zhou, and T. Yoshioka,
\newblock ``Joint speaker counting, speech recognition, and speaker
  identification for overlapped speech of any number of speakers,''
\newblock in {\em Interspeech}, 2020, pp. 36--40.

\bibitem{LibriParty}
M. Ravanelli,
\newblock ``Libriparty,''
  \url{https://github.com/speechbrain/speechbrain/tree/develop/recipes/LibriParty/generate_dataset},
  2023,
\newblock GitHub repository.

\bibitem{chung2019voxsrc}
J.~S. Chung, A. Nagrani, E. Coto, W. Xie, M. McLaren, D.~A. Reynolds, and A.
  Zisserman,
\newblock ``{VoxSRC} 2019: The first {VoxCeleb} speaker recognition
  challenge,''
\newblock {\em arXiv preprint arXiv:1912.02522}, 2019.

\bibitem{nagrani2020voxsrc}
A. Nagrani, J. Son~Chung, J. Huh, A. Brown, E. Coto, W. Xie, M. McLaren, D.~A.
  Reynolds, and A. Zisserman,
\newblock ``{VoxSRC} 2020: The second {VoxCeleb} speaker recognition
  challenge,''
\newblock {\em arXiv e-prints}, pp. arXiv--2012, 2020.

\bibitem{kudo2018sentencepiece}
T. Kudo and J. Richardson,
\newblock ``Sentencepiece: A simple and language independent subword tokenizer
  and detokenizer for neural text processing,''
\newblock in {\em 2018 Conference on Empirical Methods in Natural Language
  Processing: System Demonstrations}, 2018, pp. 66--71.

\bibitem{sctk}
NIST,
\newblock ``{SCTK},'' \url{https://github.com/usnistgov/SCTK.git}, 2024,
\newblock GitHub repository.

\end{thebibliography}

%

\end{document}